\documentclass[fleqn, paper=a4, fontsize=11pt, twocolumn]{wlscirep}
\usepackage[utf8]{inputenc}
\usepackage[T1]{fontenc}

\usepackage{graphicx}
\usepackage{caption}
\usepackage{subfigure}
\usepackage{amsmath}
\usepackage{amsfonts}
\usepackage{multirow}
\usepackage{mathtools}

\newcommand\etal{\emph{et al.}}

\title{A Boost in Revealing Subtle Facial Expressions: A Consolidated Eulerian Framework} 

\author[*]{Wei Peng}
\author[*]{Xiaopeng Hong}
\author[ ]{Yingyue Xu}
\author[+]{Guoying Zhao}

\affil[+]{corresponding author}

\affil[*]{these authors contributed equally to this work}

\begin{abstract}
Facial Micro-expression Recognition (MER) distinguishes the underlying emotional states of spontaneous subtle facial expressions. Automatic MER is challenging because that 1) the intensity of subtle facial muscle movement is extremely low and 2) the duration of ME is transient.

Recent works adopt motion magnification or time interpolation to resolve these issues. Nevertheless, {existing works divide them into two separate modules due to their non-linearity. Though such operation eases the difficulty in implementation,} it ignores their underlying connections and thus results in inevitable losses in both accuracy and speed. Instead, in this paper, we explore their underlying joint formulations and propose a consolidated Eulerian framework to reveal the subtle facial movements. It expands the temporal duration and amplifies the muscle movements in micro-expressions simultaneously. Compared to existing approaches, the proposed method can not only process ME clips more efficiently but also make subtle ME movements more distinguishable. Experiments on two public MER databases indicate that our model outperforms the state-of-the-art in both speed and accuracy.
\end{abstract}
\begin{document}

\flushbottom
\maketitle

\thispagestyle{empty}

\section*{Introduction}
\let\thefootnote\relax\footnotetext{ W.Peng, X. Hong, Y. Xu, and G. Zhao are with Center for Machine Vision and Signal Analysis, University of Oulu, Finland

This work has been accepted by 2019 14th IEEE International Conference on Automatic Face \& Gesture Recognition FG 2019.

\textcolor{blue}{Copyright @ 2019 IEEE. Personal use of this material is
permitted. However, permission to use this material for any
other purposes must be obtained from the IEEE.}}

{M}{icro}-expression recognition (MER) is an emerging facial affective analysis task to distinguish the subtle spontaneous facial expressions. Automatic MER has many potential applications like clinical diagnosis, lie detection, and human computer interaction (HCI).  MER is challenging since people tend to hide their emotions in case of being found. Studies of ME in psychology indicate that even a trained human can only achieve a MER accuracy lower than $50\%$ \cite{frank2009see}. 

Promising progresses have been made over the past decade, in  aspects such as dataset construction \cite{li2013spontaneous,yan2014casme, davison2018samm}, spatial-temporal descriptors either by hand-crafting \cite{dalal2005histograms,zhao2007dynamic, li2017towards,zong2018learning,hong2016lbp,xiaohua2017discriminative,hong2017micro,zong2018cross}, or deep learning \cite{patel2016selective,tian2018sparse,li2018can,xia2018spontaneous, xia2019spatiotemporal}, \cite{wei2018unsupervised,Tran2019Dense}, etc.     However, it is still far away from satisfaction. The difficulties are two-fold : 1). The ME related muscle movements are remarkably subtle. ME, unlike general facial expressions, only covers a small area of face so that the movements are too imperceptible to be detected. 
2). ME duration is quite short. Generally, the temporal duration of ME only lasts approximately from 1/25 to 1/3 seconds \cite{polikovsky2009facial}. 

Recent works tend to employ motion magnification or time interpolation as pre-processing steps for MER. Motion MAGnification (MAG) \cite{liu2005motion,wu2012eulerian,zhang2017video} amplifies the intensity of motion and makes subtle motion much easier to observe. Li \etal~\cite{li2017towards} amplified the subtle facial movements by employing MAG before extracting features and improved the accuracy by around $10\%$ on the CASMEII \cite{yan2014casme} database. Obvious improvements are also observed in \cite{park2009subtle, liu2016main, park2015subtle} as a result of introducing MAG. Besides MAG, time interpolation also attracts increasing attention. Pfister \etal~\cite{pfister2011recognising} proposed to use Time Interpolation Model (TIM) \cite{zhou2011towards,zhou2014compact} for MER problem, which allows to obtain sufficient frames for feature extraction even for very short expressions. Methods \cite{ li2013spontaneous,yan2014casme,li2015reading,liong2014subtle,li2017towards,huang2016spontaneous,huangdiscriminative,huang2017spontaneous} also followed a similar mechanism and got promising results.

Despite these progresses, MAG and TIM are usually treated as two separate modules \cite{yan2014casme, huang2017spontaneous,li2017towards}. Such assumption eases the difficulty in implementation. However, there are several problems. Firstly, it is prone to having the diluted problem\footnote{A phenomenon that the facial muscle movements fall into decline.} \cite{li2015reading} into the succeeding process. Secondly, the calculation of intermediate products increases the computational costs. Finally, what makes it worse, ignoring the underlying connections will inevitably result in the loss of recognition accuracy.


To address these issues, we explore the underlying joint formulations for MAG and TIM and propose a consolidated framework for revealing the subtle spatial-temporal information in ME clips.The proposed model, which is called ME-Booster, can accomplish temporal interpolating and facial muscle motion magnification simultaneously. Compared to the traditional MER systems, our ME-Booster not only avoids unnecessary separation of the two modules but also eliminates the side effect brought by the intermediate process.
We investigate our framework on two popular spontaneous MER databases: SMIC \cite{li2013spontaneous} and CASMEII \cite{yan2014casme}. Experimental results indicate that our model is computationally economic and substantially outperforms the state-of-the-art approaches.

The contributions of this paper are two-fold: 1. We propose a consolidated video motion revealing model for MER, where TIM and MAG are, for the first time, jointly formulated. More importantly, ME-Booster is a distillation of the two modules, beyond a simple knockoff just putting them together; 2. We provide fast implementations of MAG and TIM methods by exploring their underlying linear forms. It is thus computationally efficient.


\begin{figure}[thbp!]
\centering

\includegraphics[width=0.4\textwidth]{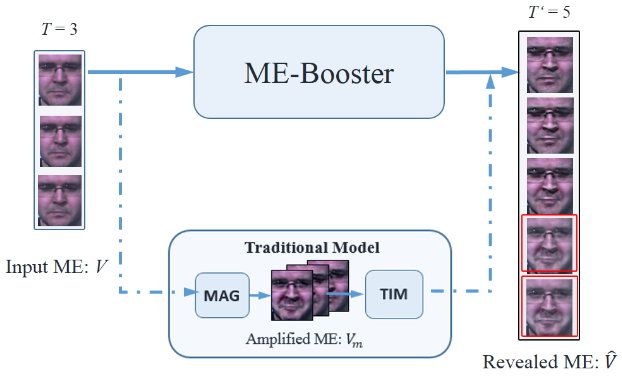}
\caption{\small{Overview of the proposed ME-Booster. The original three-frame input video can hardly be distinguished, while ME-Booster (top) expands it to five frames and makes it much easier to recognize by eyes. The traditional model (bottom) treats MAG and TIM as two parts. Note there is an intermediate product: the amplified ME without temporal extension. On the contrary, our ME-Booster model  treats these processes as an entirety. It avoids unnecessary separation of the two modules and eliminates the side effects brought by the intermediate process.}}
\label{Fig:model}
\vspace{-5mm}
\end{figure}

\section{Methodology}

In this section, we describe a consolidated Eulerian subtle motion revealing model, termed ME-Booster, to boost micro-expression recognition. Suppose there is a ME video clip $V = \left ( I(1),I(2),\dots,I(T) \right ) $, where $I(t) \in \mathbb{R}^{d}$ denotes the $t$-th frame of the video for $t = 1,\cdots,T$. Our goal is to produce a video $\hat{V}$ where the micro-expression can be easily observed. 



 The model, in its general form, is formulated by a video-specific function $f$ with a set of parameters $ \Theta$ as follows:
\begin{equation} 
\label{eq:eq_general}
  \hat{V} = f(V; \Theta),
\end{equation}
where the function $f: \mathbb{R}^{d\times T} \mapsto \mathbb{R}^{d\times {T}'}$ maps the input video to a fixed length ($T'$) one of which the subtle facial muscle movements in $V$ are revealed. 

{Considering the non-linear nature of the representative motion magnification and temporal interpolation methods, such as the Eulerian based MAG method \cite{wu2012eulerian} and the graph spectral TIM method~\cite{burt1987laplacian}, existing MER approaches \cite{yan2014casme, huang2017spontaneous,li2017towards} usually}
treat MAG and TIM as two separate modules and adopt them in {a successive manner to alleviate the difficulty in implementation. As discussed above, it may easily lead to losses in both accuracy and speed.} 

{To the contrary, in this paper} we will show that under {certain but reasonable circumstances}, MAG and TIM can be embedded in Eq.~(\ref{eq:eq_general}) by a scale-wise linear function.
Firstly, the Eulerian MAG model \cite{liu2005motion,wu2012eulerian,zhang2017video} is  multi-scale and non-linear, and thus a complicated system. However, we realize that it is feasible to implement it as a series of scale-wise linear functions, since in each of the scales it approximates movements using the Taylor expansion.  
Secondly, 
we will show that the {interpolated video can be obtained by multiplying the input video with a matrix, without explicitly solving the generalized eignvalue problem as the graph spectral TIM method \cite{zhou2011towards} does. 
Thus, w.l.o.g., in the following context, we formulate our model with a `single-resolution' setting and consider the $f(\cdot)$ as a linear function:\footnote{\emph{Multi-resolution analysis} can be easily reached by applying the model to each single resolution.}}

\begin{equation} \label{eq:all}
  \hat{V} =V\mathbf{W}.
\end{equation}
To obtain  the parameter matrix $ \mathbf{W} \in \mathbb{R}^{T\times T^{'}} $, we decouple it as follow: 
\begin{equation}\label{eq:orig}
  \mathbf{W} = \mathbf{W}^{M}\mathbf{W}^{I},
\end{equation}
in which $\mathbf{W}^{M}\in \mathbb{R}^{T \times T} $ and $\mathbf{W}^{I} \in \mathbb{R}^{T\times T^{'}}$ are the magnification matrix and the interpolation matrix, respectively.  The order of the right hand side of Eq. (\ref{eq:orig}) is consistent with the one in \cite{li2017towards}. 

In the following paragraphs, we will derive the solutions of $\mathbf{W}^{M}$ and $\mathbf{W}^{I}$ respectively in detail {and show that both $\mathbf{W}^{M}$ and $\mathbf{W}^{I}$ can be determined once the lengths of the input and output video are given. As a result, Eqs. \ref{eq:all} and \ref{eq:orig} ensure that $\mathbf{W}$ is obtained before being applied on the input video in practice.}

\vskip 0.1in
\textbf{Solution to the magnification matrix $\mathbf{W}^{M}$}

In this subsection, we aim to get the solution for $\mathbf{W}^{M}$.
Let $W^{M}_{i,j}$ be the entry of $\mathbf{W}^{M}$ at $\left (i, j \right )$ for $i,j = 1,\cdots,T$. 
The magnified frame $I_m(t)$ can be described as :
\begin{equation}\label{eq:amp}
   I_m(t) = \sum_{i=1}^{T} I(i)W^{M}_{i,t}.
\end{equation}
To get the solution of $\mathbf{W}^{M}$, we revisit the Eulerian magnification model \cite{wu2012eulerian}. It regards the first frame in a video as a digital signal $s(x)$, and every frame as a displacement of this signal, i.e., $I(x,t) = s(x+\delta(t))$, where $I(x,t)$ is the pixel intensity at location $x$ of the $t$-th frame and $\delta(t)$ is a tiny displacement function. It is then approximated by the
first-order Taylor expansion : 
\begin{equation}\label{eq:sig}
s(x+\delta(t)) = s(x) + \delta(t) \frac{\partial s}{\partial x},
\end{equation}
where every frame in the ME video is 
an addition of the base signal and the multiplication of the displacement and the first derivative. 
 Here, $\frac{\partial s}{\partial x} = I(t)-I(t-1)$ represents the ME motion. 
 It is fulfilled using a recursive algorithm to take all the preceding frames into account and get the `historical' motion $\mathbf{B}(t)$:
\begin{equation}\label{eq:filter}
   \mathbf{B}(t) = L_1(t) - L_2(t).
\end{equation}
$L_1(t)$ and $L_2(t)$, for $t = 2,\cdots,T$ are auxiliary variables:
\begin{equation}\label{eq:lowpass}
\begin{cases}
L_1(t) = w_1I(t) + (1-w_1)L_1(t-1)\\
L_2(t) = w_2I(t) + (1-w_2)L_2(t-1),\\ 
\end{cases}
\end{equation}
where the $L_1(1)= I(1),~L_2(1)= I(1)$. The hyper-parameters $w_1$ and $w_2$ are in the range $(0,1)$ and $w_1 > w_2$. Then applying a magnification factor $\alpha$ to $\mathbf{B}(t)$, we get the magnified frame $I_m(t)$ :
\begin{equation}\label{eq:amplify}
   I_m(t) = I(t) + \alpha \mathbf{B}(t).
\end{equation}

The Eulearian magnification model in  Eq. (\ref{eq:amplify}) is clearly a linear model. We solve out a parameter matrix operated on the input video by expanding the recursive form in (\ref{eq:lowpass}). As a result, by combining Eq. (\ref{eq:amp}) and Eq. (\ref{eq:amplify}), we get the solution for $\mathbf{W}^M$:
\begin{equation}\label{eq:MAG}
W^{M}_{i,j} = \begin{cases}
\alpha (1-w_1)^a w_1^b - \alpha(1-w_2)^a w_2^b &\text{if $j > i$ }\\
\alpha(w_1 -w_2) + 1  &\text{if $j = i$}\\
0 &\text{else},
\end{cases}
\end{equation}
where $a={j-i}$ and $b = {min(1,i-1)}$. As it is in Eq. (\ref{eq:MAG}), $\mathbf{W}^{M}$ is an upper triangular matrix since ME motion is only related to the preceding video frames.
 Once the lengths of the input and output videos are determined, the entry of $\mathbf{W}^{M}$ can be computed and stored by a look-up table (LUT). That will further improve the computational efficiency.
To get a better magnification result, one can employ $\mathbf{W}^M$ on the multi-scale input video, where for different scales, the factor $\alpha$ in Eq. (\ref{eq:MAG}) may be truncated.
 
\vskip 0.1in
\textbf{Solution to the interpolation matrix $\mathbf{W}^{I}$ }

In this subsection, we will derive $\mathbf{W}^{I}$ in Eq. (\ref{eq:orig}) to form our efficient time interpolation module for extending $V_m$ to $T'$ frames. Just for clarification, we introduce the symbols $V_m = \{I_{m}(t)\}$ for $t = 1,2,\cdots, T$  to describe the intermediate magnification output, though they are finally eliminated by Eq.~\ref{eq:orig} in practice.
The recovered video $\hat{V}$ is:
\begin{equation}\label{eq:ti}
  \hat{V} =V_m \mathbf{W}^{I}.
\end{equation}
Applying a graph spectral method \cite{burt1987laplacian}, the traditional TIM approach \cite{zhou2011towards} finds projection functions by which TIM keeps the frame adjacency priority in the low latent space. That is
\begin{equation}\label{eq:map}
  \mathcal{F}_{map}(I_m(t)) = (\mathbf{M}^{-1}\mathcal{V}^\top \mathbf{U}^\top)(I_m(t)-\Bar{I_m}), 
\end{equation}
where the function $\mathcal{F}_{map}: \mathbb{R}^{d} \mapsto \mathbb{R}^{T-1}$ describes a projection from an image to $T-1$ points. Here, $\mathbf{U}$ is an unitary matrix from SVD~\cite{golub2012matrix} on input, $\mathbf{M}$ and $\mathcal{V}$ are graph Laplacian related diagonal matrix and eigenvectors, respectively. $\Bar{I_m}$ is the average frame of all the frames in video $V_m$. 
Readers may refer to \cite{zhou2011towards} for more details.

In this way, every frame becomes a group of discrete points in a latent space. According to \cite{burt1987laplacian}, the optimal projection is a group of sine {functions. Let $\mathcal{F}(t) = \mathcal{F}_{map}(I_m(t))$.
By} sampling from the latent space and projecting the points back to image space, we reconstruct the input. According to Eq. (\ref{eq:map}), the intermediate magnification video can be reconstructed by
\begin{equation}\label{eq:imap}
V_m = \mathbf{A} \mathbf{Y}  + \Bar{V_m},
\end{equation}
where $\Bar{V_m} = V_m (\frac{1}{T}\mathbf{1}_{T \times T})$, $\mathcal{V}^{-1}$ exists since $\mathcal{V}$ is full rank. So $\mathbf{A} = \mathbf{U}(\mathcal{V}^{-1})^\top\mathbf{M}$, and $\mathbf{Y} = [\mathcal{F}(\frac{1}{T}),\mathcal{F}(\frac{2}{T}),...,\mathcal{F}(1)]$ are $T$ groups of points in latent space. Each $\mathcal{F}(t)$, which are sampled from $T-1$ sine functions, refers to one frame. So sampling $T^{'}$ groups of points, say $\mathbf{Y}^{'} = [\mathcal{F}(\frac{1}{T^{'}}), \mathcal{F}(\frac{2}{T^{'}}), ... ,\mathcal{F}(1)]$ , the video is extended to $T^{'}$ frames. That is :
\begin{equation}\label{eq:imapV}
\hat{V} = \mathbf{A} \mathbf{Y'}  + \Bar{V_m'},
\end{equation}
 where $\Bar{V_m'}(t) = V_m (\frac{1}{T}\mathbf{1}_{T \times T'})$. Since $\mathbf{Y'}$ is a group of discrete points, it satisfies the requirement for a linear TIM model. 
 According to Eq. (\ref{eq:imap}), we adopt the right inverse matrix to compute $\mathbf{A}$ :
\begin{equation}\label{eq:inverse}
  \mathbf{A} = V_m(\mathbf{I}-\frac{1}{T}\mathbf{1}_{T \times T})\mathbf{Y}^\top (\mathbf{Y} \mathbf{Y}^\top)^{-1}.
\end{equation}
Substituting Eq. (\ref{eq:inverse}) into Eq. (\ref{eq:imapV}), we have
\begin{equation}\label{eq:interpolation} 
\hat{V} = V_m(\mathbf{I}-\frac{1}{T}\mathbf{1}_{T \times T})\mathbf{Y}^\top (\mathbf{Y} \mathbf{Y}^\top)^{-1}\mathbf{Y}^{'} +\Bar{V_m'}.
\end{equation}
Finally we obtain the solution for $\mathbf{W}^{I}$, that is:
\begin{equation}\label{eq:WI} 
\mathbf{W}^{I} = (\mathbf{I}-\frac{1}{T}\mathbf{1}_{T \times T})\mathbf{Y}^\top (\mathbf{Y} \mathbf{Y}^\top)^{-1}\mathbf{Y}^{'} +\frac{1}{T}\mathbf{1}_{T \times T'}.
\end{equation}

Here Eq. (\ref{eq:WI}) provides us a linear solution to $\mathbf{W}^{I}$ and it is independent to the input video, which means $\mathbf{W}^{I}$ also can be computed in advance. 

After obtaining $\mathbf{W}^{M}$ and $\mathbf{W}^{I}$, we employ Eq. (\ref{eq:orig}) to solve the matrix $\mathbf{W}$, which is the only parameter matrix used in practice.  Finally, a consolidated model is constructed by using this linear operator $\mathbf{W}$. Obviously, ME-booster is a unique model for both video magnification and time interpolation. The relation to the magnification and interpolation method is clear. When $\mathbf{W}^{I}$ is an identity matrix, the model formulated in Eq. (\ref{eq:all})  is  a  MAG  model; When $\mathbf{W}^{M}$ is  an  identity matrix, it degenerates to a TIM model. 

Our ME-booster model is computationally economic and can make the subtle ME more distinguished. These advantages are mainly contributed by the discovery of the underlying common linear property between TIM and MAG modules. By this way, there are no need of intermediate process and complicated computation of graph Laplacian. Besides, expanding recursive forms and decoupling model parameters from the input also dedicate to these advantages.

\addtolength{\textheight}{-3cm}   

\section{Experiments}

We evaluate the proposed model on two widely used spontaneous ME databases: SMIC~\cite{li2013spontaneous}  and CASMEII~\cite{yan2014casme}. The SMIC database contains 306 ME video clips belonging to four sub-sets, which are SMIC-HS, SMIC-VIS, SMIC-NIR, and SMIC-subHS. There are three classes for all the ME clips : \emph{Positive, Negative}, and \emph{Surprise}.  The CASMEII database contains 247 ME video clips. The baseline method of CASMEII chooses classes \emph{Happiness, Surprise, Disgust, Repression, }and \emph{Others} for training and testing.



\begin{figure*}[thbp!]
    \centering
    \subfigure[HIGO]{
        \label{fig:HIGO}    
        \includegraphics[width=0.29\textwidth]{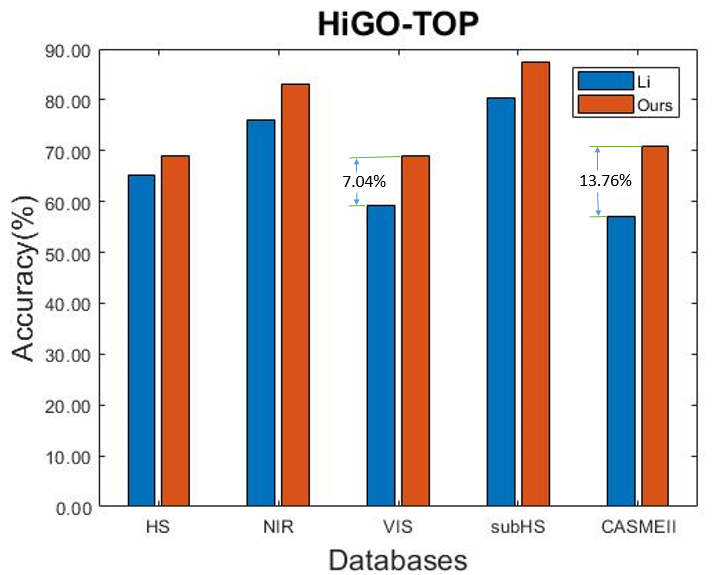}
    }    
    \subfigure[HOG]{
        \label{fig:HOG}    
        \includegraphics[width=0.29\textwidth]{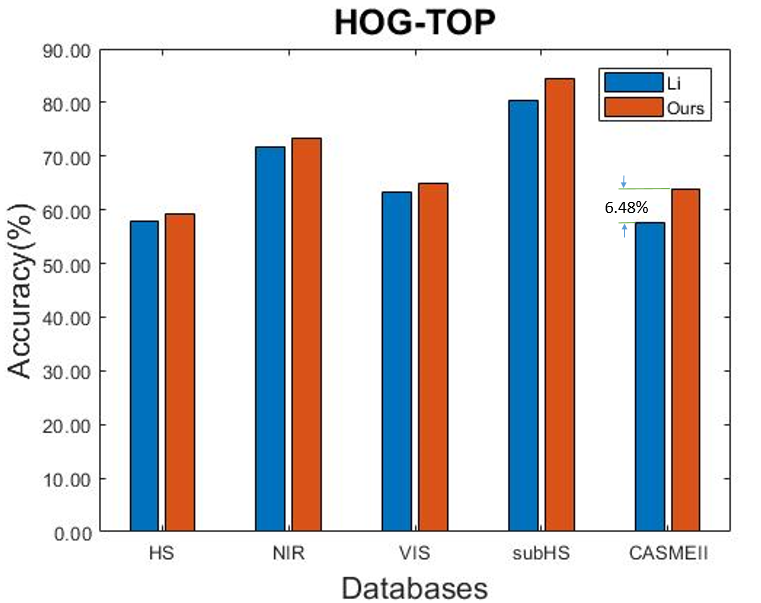}
    }  
    \subfigure[LBP]{
        \label{fig:LBP}    
        \includegraphics[width=0.29\textwidth]{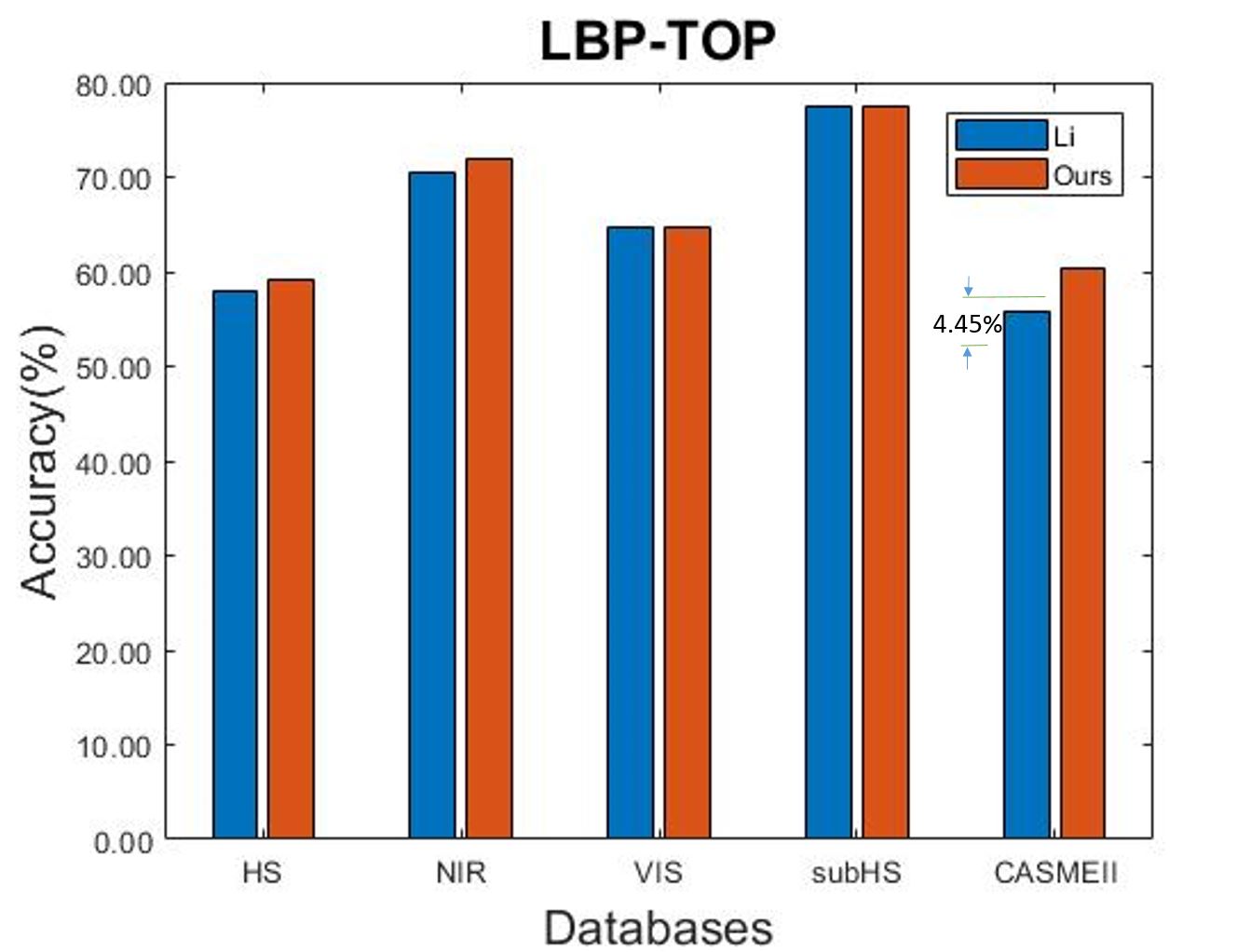}
    }
    \caption {\small{MER performance comparison using three mainstream feature descriptors: (a) HIGO-TOP, (b) HOG-TOP, and (c) LBP-TOP.}}
    \label{fig:MER 3D}
    \vspace{-6mm}
\end{figure*}




To extensively evaluate the proposed ME-Booster, we perform the following three experiments: A). Consolidation versus separate; B). Comparisons to the state of the arts; and C). Computational cost evaluation.{We explore the different values for hyper-parameters $T'$, $\alpha$, and $w_1$ and $w_2$. Based on the results, which are found consistent to those in \cite{li2017towards, wu2012eulerian}, the parameters are empirically} set to 10, 16, 0.4, and 0.05, respectively.
For the experiments A) and B),
a linear SVM \cite{burges1998tutorial} is used as classifier for our model, unless further specified. As suggested by \cite{li2017towards}, in all MER experiments, the Leave-One-Subject-Out (LOSO) protocol is adopted. The micro-average method \cite{van2013macro} is employed as evaluation metrics.

\textbf{A. \emph{Consolidation} versus \emph{separate}}

We compare ME-Booster with the current best method \cite{li2017towards}, which contains both the MAG and TIM modules as two separate ones, as illustrated as the 'traditional model' in Fig. \ref{Fig:model}.
For fair comparison, we use the same classifier and three mainstream temporal-spatial feature descriptors: LBP-TOP\cite{zhao2007dynamic}, HOG-TOP\cite{dalal2005histograms} and HIGO-TOP \cite{li2017towards}, which are extensively evaluated in the state of the art \cite{li2017towards}.


The comparative results are reported in Fig. \ref{fig:MER 3D}. There are several important observations. 1) For HIGO-TOP, the accuracy is improved by almost 10.0\% on the SMIC-NIR. On the SMIC-subHS, our accuracy is 87.32\% , while the current best is only 80.28\%. 
2) For the HOG-TOP, our model improves the accuracy by 3.0\% on average.
3) For the LBP-TOP, ME-Booster is slightly better than the baseline on the SIMC, while it improves by almost 4.5\% on the CASMEII.
4) The largest improvements in accuracy are almost 13\%, 6\%, and 4.5\% for the HIGO-TOP, HOG-TOP, and LBP-TOP respectively. From the results, it is safe to conclude that our model achieves superior accuracy to the current best \cite{li2017towards}, regardless of the video descriptors chosen.






\textbf{B. Comparison to the state of the arts}

We compare our method to nine state-of-the-art approaches, including both the deep learning method \cite{patel2016selective} and the conventional hand-crafting feature approaches \cite{li2013spontaneous,yan2014casme, park2015subtle,wang2015micro, liong2014optical,huangdiscriminative, huang2017spontaneous,li2017towards}. The current best method \cite{li2017towards} provides a standard pipeline for MER tasks.
As the HIGO-TOP descriptor is recommended by~\cite{li2017towards}, we also use it as the video descriptors. 
 The results are summarized in Table. \ref{table:MER_final} and Table. \ref{table:MER_sub}. For all the databases, our method achieves the best accuracy against all the other methods. More specifically, the accuracy on SMIC-HS, SMIC-subHS, SMIC-VIS, SMIC-NIR, and CASMEII are 68.90\%, 87.32\%, 83.10\%, 69.01\%, and 70.85\% respectively, while the second best results achieved by \cite{li2017towards} are only 68.29\%, 80.28\%, 81.69\%, 67.61\%, and 67.21\% respectively.
 It indicates that our ME-Booster is robust to the substantial variations in appearance brought by different sensors. 

\begin{table}
\centering
\caption{\small{Experimental results of MER comparing to recent  state-of-the-art approaches on the SMIC and the CASMEII databases. }}
\scalebox{0.8}{
 \begin{tabular}{c| c| c}
 \hline
   &SMIC-HS & CASMEII \\ 
 \hline\hline
Baseline-SMIC \cite{li2013spontaneous} & 48.8\% & - \\ 
 \hline
 Baseline-CASMEII \cite{yan2014casme} & - & 63.41\% \\
 \hline\hline
  Adaptive Mag \cite{park2015subtle} & - & 51.9\% \\
 \hline
 Wang \etal \cite{wang2015micro} & - & 62.3\% \\
 \hline
 Liong \etal\cite{liong2014optical} & 53.56\% & - \\
 \hline
 Huang \etal\cite{huang2017spontaneous} & 60.98\% &64.37\% \\ 
 \hline
 DiSTLBP-RIP \cite{huangdiscriminative} & 63.41\% & 64.78\% \\ 
 \hline
   Patel \etal\cite{patel2016selective} & 53.60\% & 47.30\% \\ 
 \hline
  Li \etal\cite{li2017towards} & 68.29\%& 67.21\% \\ 
 \hline
 \textbf{ Ours} & \textbf{68.90\%} & \textbf{70.85\%}\\  
 \hline
\end{tabular}}
\label{table:MER_final}
\end{table}

\begin{table}
\centering
\caption{\small{Experimental results of MER on sub-datasets of SMIC.}}
\scalebox{0.8}{
 \begin{tabular}{c| c| c |c}
 \hline
   &SMIC-subHS & SMIC-VIS & SMIC-NIR  \\ 
 \hline\hline
Baseline-SMIC \cite{li2013spontaneous} & - & 52.1\%& 38.0\% \\ 
 \hline\hline
  Li \etal\cite{li2017towards}  & 80.28\% & 81.69\%& 67.61\% \\ 
 \hline
 \textbf{ Ours} &\textbf{87.32\%} & \textbf{83.10\%} & \textbf{69.01\%}\\  
 \hline
\end{tabular}}
\label{table:MER_sub}
\vspace{-5mm}
\end{table}

\textbf{C. Computational cost evaluation}

Finally, we compare the computational cost of ME-Booster to the current best work~\cite{li2017towards}. The experiment is conducted on MATLAB R2017b platform, and the desktop is with an Intel CPU i5-6500 (3.20GHz) and RAM of 8 GB.  Results indicate that our model is more efficient on both databases. For SMIC, the running time are 125.49 seconds (s) for Li \etal~\cite{li2017towards} and 27.42 s for our model. For CASMEII, our model can be even ten times faster than the model in \cite{li2017towards}, the running time are 420.89 s and 41.61 s, respectively.

 

\section{CONCLUSIONS}

 We explore the underlying relationship between MAG and TIM and reach their common linear nature. In line with this important finding, we propose a consolidated Eulerian framework, ME-Booster, for revealing the subtle and fleeting facial expressions. ME-Booster boosts the performance of MER in both speed and accuracy. On one hand, it avoids unnecessary separation of the two modules and thus the computational cost is reduced. On the other hand, the accuracy is further improved since it eliminates the side effects brought by the intermediate process. ME-Booster is thus a distillation of the MAG and TIM, beyond a simple knockoff just putting them together. It is online, training-free, and compatible with any succeeding processing methods in the MER pipeline. Experiments on two databases indicate that our model outperforms the state-of-the-art approaches in an efficient manner.

\section{ACKNOWLEDGMENTS}

This work was supported by the Academy of Finland, Tekes Fidipro program (Grant No. 1849/31/2015), Business Finland project (Grant No. 3116/31/2017), and Infotech Oulu. It was also supported, in part, by the National Natural Science Foundation of China (No. 61772419 $\&$ 61572205). Moreover, the authors wish to acknowledge CSC-IT Center for Science, Finland, for computational resources. Furthermore, we express deep gratitude to the support of NVIDIA Corporation with the donation of GPUs for this research.

\bibliographystyle{ieeetr}
\bibliography{sample}

\begin{thebibliography}{10}

\bibitem{frank2009see}
M.~Frank, M.~Herbasz, K.~Sinuk, A.~Keller, and C.~Nolan, ``I see how you feel:
  Training laypeople and professionals to recognize fleeting emotions,'' in
  {\em The Annual Meeting of the International Communication Association.
  Sheraton New York, New York City}, 2009.

\bibitem{li2013spontaneous}
X.~Li, T.~Pfister, X.~Huang, G.~Zhao, and M.~Pietik{\"a}inen, ``A spontaneous
  micro-expression database: Inducement, collection and baseline,'' in {\em
  Automatic face and gesture recognition (FG), 2013 10th IEEE International
  Conference on}, pp.~1--6, IEEE, 2013.

\bibitem{yan2014casme}
W.-J. Yan, X.~Li, S.-J. Wang, G.~Zhao, Y.-J. Liu, Y.-H. Chen, and X.~Fu,
  ``Casme ii: An improved spontaneous micro-expression database and the
  baseline evaluation,'' {\em PloS one}, vol.~9, no.~1, 2014.

\bibitem{davison2018samm}
A.~K. Davison, C.~Lansley, N.~Costen, K.~Tan, and M.~H. Yap, ``Samm: A
  spontaneous micro-facial movement dataset,'' {\em IEEE Transactions on
  Affective Computing}, vol.~9, no.~1, pp.~116--129, 2018.

\bibitem{dalal2005histograms}
N.~Dalal and B.~Triggs, ``Histograms of oriented gradients for human
  detection,'' in {\em Computer Vision and Pattern Recognition(CVPR).2005. IEEE
  Conference on}, vol.~1, pp.~886--893, IEEE, 2005.

\bibitem{zhao2007dynamic}
G.~Zhao and M.~Pietikainen, ``Dynamic texture recognition using local binary
  patterns with an application to facial expressions,'' {\em IEEE transactions
  on pattern analysis and machine intelligence (TPAMI)}, vol.~29, no.~6,
  pp.~915--928, 2007.

\bibitem{li2017towards}
X.~Li, X.~Hong, A.~Moilanen, X.~Huang, T.~Pfister, G.~Zhao, and M.~Pietikainen,
  ``Towards reading hidden emotions: A comparative study of spontaneous
  micro-expression spotting and recognition methods,'' {\em IEEE Transactions
  on Affective Computing}, 2017.

\bibitem{zong2018learning}
Y.~Zong, X.~Huang, W.~Zheng, Z.~Cui, and G.~Zhao, ``Learning from hierarchical
  spatiotemporal descriptors for micro-expression recognition,'' {\em IEEE
  Transactions on Multimedia}, 2018.

\bibitem{hong2016lbp}
X.~Hong, Y.~Xu, and G.~Zhao, ``Lbp-top: a tensor unfolding revisit,'' in {\em
  Asian Conference on Computer Vision}, pp.~513--527, Springer, 2016.

\bibitem{xiaohua2017discriminative}
H.~Xiaohua, S.-J. Wang, X.~Liu, G.~Zhao, X.~Feng, and M.~Pietikainen,
  ``Discriminative spatiotemporal local binary pattern with revisited integral
  projection for spontaneous facial micro-expression recognition,'' {\em IEEE
  Transactions on Affective Computing}, 2017.

\bibitem{hong2017micro}
X.~Hong, T.-K. Tran, and G.~Zhao, ``Micro-expression spotting: A benchmark,''
  {\em arXiv preprint arXiv:1710.02820}, 2017.

\bibitem{zong2018cross}
Y.~Zong, W.~Zheng, X.~Hong, C.~Tang, Z.~Cui, and G.~Zhao, ``Cross-database
  micro-expression recognition: A benchmark,'' {\em arXiv preprint
  arXiv:1812.07742}, 2018.

\bibitem{patel2016selective}
D.~Patel, X.~Hong, and G.~Zhao, ``Selective deep features for micro-expression
  recognition,'' in {\em Pattern Recognition (ICPR), 2016 23rd International
  Conference on}, pp.~2258--2263, IEEE, 2016.

\bibitem{tian2018sparse}
L.~Tian, X.~Hong, C.~Fan, Y.~Ming, M.~Pietik{\"a}inen, and G.~Zhao, ``Sparse
  tikhonov-regularized hashing for multi-modal learning,'' in {\em 2018 25th
  IEEE International Conference on Image Processing (ICIP)}, pp.~3793--3797,
  IEEE, 2018.

\bibitem{li2018can}
Y.~Li, X.~Huang, and G.~Zhao, ``Can micro-expression be recognized based on
  single apex frame?,'' in {\em 2018 25th IEEE International Conference on
  Image Processing (ICIP)}, pp.~3094--3098, IEEE, 2018.

\bibitem{xia2018spontaneous}
Z.~Xia, X.~Feng, X.~Hong, and G.~Zhao, ``Spontaneous facial micro-expression
  recognition via deep convolutional network,'' in {\em 2018 Eighth
  International Conference on Image Processing Theory, Tools and Applications
  (IPTA)}, pp.~1--6, IEEE, 2018.

\bibitem{xia2019spatiotemporal}
Z.~Xia, X.~Hong, X.~Gao, X.~Feng, and G.~Zhao, ``Spatiotemporal recurrent
  convolutional networks for recognizing spontaneous micro-expressions,'' {\em
  arXiv preprint arXiv:1901.04656}, 2019.

\bibitem{wei2018unsupervised}
X.~Wei, H.~Li, J.~Sun, and L.~Chen, ``Unsupervised domain adaptation with
  regularized optimal transport for multimodal 2d+ 3d facial expression
  recognition,'' in {\em Automatic Face \& Gesture Recognition (FG 2018), 2018
  13th IEEE International Conference on}, pp.~31--37, IEEE, 2018.

\bibitem{Tran2019Dense}
X.~Hong, Q.-N. Vo, T.-K. Tran, and G.~Zhao, ``Dense prediction for
  micro-expression spotting based on deep sequence model,'' 2019.

\bibitem{polikovsky2009facial}
S.~Polikovsky, Y.~Kameda, and Y.~Ohta, ``Facial micro-expressions recognition
  using high speed camera and 3d-gradient descriptor,'' 2009.

\bibitem{liu2005motion}
C.~Liu, A.~Torralba, W.~T. Freeman, F.~Durand, and E.~H. Adelson, ``Motion
  magnification,'' {\em ACM transactions on graphics (TOG)}, vol.~24, no.~3,
  pp.~519--526, 2005.

\bibitem{wu2012eulerian}
H.~Wu, M.~Rubinstein, E.~Shih, J.~Guttag, F.~Durand, and W.~T. Freeman,
  ``Eulerian video magnification for revealing subtle changes in the world,''
  {\em ACM Trans. Graph. (Proceedings SIGGRAPH 2012)}, vol.~31, no.~4, 2012.

\bibitem{zhang2017video}
Y.~Zhang, S.~L. Pintea, and J.~C. van Gemert, ``Video acceleration
  magnification,'' {\em arXiv preprint arXiv:1704.04186}, 2017.

\bibitem{park2009subtle}
S.~Park and D.~Kim, ``Subtle facial expression recognition using motion
  magnification,'' {\em Pattern Recognition Letters}, vol.~30, no.~7,
  pp.~708--716, 2009.

\bibitem{liu2016main}
Y.-J. Liu, J.-K. Zhang, W.-J. Yan, S.-J. Wang, G.~Zhao, and X.~Fu, ``A main
  directional mean optical flow feature for spontaneous micro-expression
  recognition,'' {\em IEEE Transactions on Affective Computing}, vol.~7, no.~4,
  pp.~299--310, 2016.

\bibitem{park2015subtle}
S.~Y. Park, S.~H. Lee, and Y.~M. Ro, ``Subtle facial expression recognition
  using adaptive magnification of discriminative facial motion,'' in {\em
  Proceedings of the 23rd ACM international conference on Multimedia},
  pp.~911--914, ACM, 2015.

\bibitem{pfister2011recognising}
T.~Pfister, X.~Li, G.~Zhao, and M.~Pietik{\"a}inen, ``Recognising spontaneous
  facial micro-expressions,'' in {\em Computer Vision (ICCV), 2011 IEEE
  International Conference on}, pp.~1449--1456, IEEE, 2011.

\bibitem{zhou2011towards}
Z.~Zhou, G.~Zhao, and M.~Pietik{\"a}inen, ``Towards a practical lipreading
  system,'' in {\em Computer Vision and Pattern Recognition (CVPR), 2011 IEEE
  Conference on}, pp.~137--144, IEEE, 2011.

\bibitem{zhou2014compact}
Z.~Zhou, X.~Hong, G.~Zhao, and M.~Pietik{\"a}inen, ``A compact representation
  of visual speech data using latent variables,'' {\em IEEE transactions on
  pattern analysis and machine intelligence}, vol.~36, no.~1, pp.~1--1, 2014.

\bibitem{li2015reading}
X.~Li, X.~Hong, A.~Moilanen, X.~Huang, T.~Pfister, G.~Zhao, and
  M.~Pietik{\"a}inen, ``Reading hidden emotions: spontaneous micro-expression
  spotting and recognition,'' {\em arXiv preprint arXiv:1511.00423}, 2015.

\bibitem{liong2014subtle}
S.-T. Liong, J.~See, R.~C.-W. Phan, A.~C. Le~Ngo, Y.-H. Oh, and K.~Wong,
  ``Subtle expression recognition using optical strain weighted features,'' in
  {\em Asian Conference on Computer Vision}, pp.~644--657, Springer, 2014.

\bibitem{huang2016spontaneous}
X.~Huang, G.~Zhao, X.~Hong, W.~Zheng, and M.~Pietik{\"a}inen, ``Spontaneous
  facial micro-expression analysis using spatiotemporal completed local
  quantized patterns,'' {\em Neurocomputing}, vol.~175, pp.~564--578, 2016.

\bibitem{huangdiscriminative}
X.~Huang, S.-J. Wang, X.~Liu, G.~Zhao, X.~Feng, and M.~Pietikainen,
  ``Discriminative spatiotemporal local binary pattern with revisited integral
  projection for spontaneous facial micro-expression recognition,'' {\em IEEE
  Transactions on Affective Computing}, 2017.

\bibitem{huang2017spontaneous}
X.~Huang and G.~Zhao, ``Spontaneous facial micro-expression analysis using
  spatiotemporal local radon-based binary pattern,'' in {\em the Frontiers and
  Advances in Data Science (FADS), 2017 International Conference on},
  pp.~159--164, IEEE, 2017.

\bibitem{burt1987laplacian}
P.~J. Burt and E.~H. Adelson, ``The laplacian pyramid as a compact image
  code,'' in {\em Readings in Computer Vision}, pp.~671--679, Elsevier, 1987.

\bibitem{golub2012matrix}
G.~H. Golub and C.~F. Van~Loan, {\em Matrix computations}, vol.~3.
\newblock JHU Press, 2012.

\bibitem{burges1998tutorial}
C.~J. Burges, ``A tutorial on support vector machines for pattern
  recognition,'' {\em Data mining and knowledge discovery}, vol.~2, no.~2,
  pp.~121--167, 1998.

\bibitem{van2013macro}
V.~Van~Asch, ``Macro-and micro-averaged evaluation measures [[basic draft]],''
  2013.

\bibitem{wang2015micro}
S.-J. Wang, W.-J. Yan, X.~Li, G.~Zhao, C.-G. Zhou, X.~Fu, M.~Yang, and J.~Tao,
  ``Micro-expression recognition using color spaces,'' {\em IEEE Transactions
  on Image Processing}, vol.~24, no.~12, pp.~6034--6047, 2015.

\bibitem{liong2014optical}
S.-T. Liong, R.~C.-W. Phan, J.~See, Y.-H. Oh, and K.~Wong, ``Optical strain
  based recognition of subtle emotions,'' in {\em Intelligent Signal Processing
  and Communication Systems (ISPACS), 2014 International Symposium on},
  pp.~180--184, IEEE, 2014.

\end{thebibliography}

\end{document}